%% file: main.tex
\documentclass[journal]{IEEEtran}
\usepackage{ifpdf}
\ifCLASSINFOpdf
  \usepackage[pdftex]{graphicx}
  \usepackage{graphbox}
  \graphicspath{{figures/}}
  \DeclareGraphicsExtensions{.pdf,.jpeg,.png}
\else
  \usepackage[dvips]{graphicx}
  \graphicspath{{figures/}}
  \DeclareGraphicsExtensions{.eps}
\fi
\usepackage{amsmath}
\usepackage[bookmarks=false]{hyperref} 
\usepackage{colortbl}
\usepackage{booktabs}
\usepackage[per-mode=symbol,range-phrase=--,range-units=single]{siunitx}
\usepackage[table,dvipsnames]{xcolor}
\usepackage{cite}
\usepackage{nicefrac}
\usepackage{multirow}
\usepackage{calc}
\usepackage{printlen}
\usepackage[normalem]{ulem}

\begin{document}
\title{A Deep Active Contour Model for Delineating Glacier Calving Fronts}

\author{Konrad~Heidler,~\IEEEmembership{Student~Member,~IEEE},
        Lichao~Mou,
        Erik~Loebel,
        Mirko~Scheinert,
        Sébastien~Lefèvre,~\IEEEmembership{Senior~Member,~IEEE}
        and~Xiao~Xiang~Zhu,~\IEEEmembership{Fellow,~IEEE}%
\thanks{%
  K.~Heidler, L.~Mou and X.~Zhu are with the
  Chair of Data Science in Earth Observation (SiPEO),
  Department of Aerospace and Geodesy,
  School of Engineering and Design,
  Technical University of Munich (TUM),
  80333 Munich
  E-mails: k.heidler@tum.de; lichao.mou@tum.de; xiaoxiang.zhu@tum.de
}%
\thanks{%
  E.~Loebel and M.~Scheinert are with the
  Institut für Planetare Geodäsie, Technische Universität Dresden, 01069 Dresden, Germany.
  E-mails: erik.loebel@tu-dresden.de, mirko.scheinert@tu-dresden.de
}
\thanks{%
  S.~Lefèvre is with IRISA UMR 6074, Université Bretagne Sud, 56000 Vannes, France.
  E-mail: sebastien.lefevre@univ-ubs.fr
}
\thanks{
  A preliminary version of this study was presented at IGARSS 2022.
}
\thanks{%
  This work is supported by the Helmholtz Association through the Helmholtz Information and Data Science Incubator project ``Artificial Intelligence for Cold Regions'', Acronym \emph{AI-Core}, by Helmholtz Association’s Initiative and Networking Fund through Helmholtz AI [grant number: ZT-I-PF-5-01] -- Local Unit ``Munich Unit @Aeronautics, Space and Transport (MASTr)'', by the German Federal Ministry of Education and Research (BMBF)
  in the framework of the international future AI lab
  ``AI4EO -- Artificial Intelligence for Earth Observation: Reasoning, Uncertainties, Ethics and Beyond'' (Grant number: 01DD20001) and by German Federal Ministry for Economic Affairs and Climate Action in the framework of the "national center of excellence ML4Earth" (grant number: 50EE2201C).
  (Corresponding authors: Xiao Xiang Zhu, Lichao Mou.)
}%
}
\markboth{SUBMITTED TO IEEE TRANSACTIONS ON GEOSCIENCE AND REMOTE SENSING}%
{Heidler \MakeLowercase{\textit{et al.}}: Monitoring the Antarctic Coastline by Combining Semantic Segmentation and Edge Detection}

\maketitle
\begin{abstract}
  \textcolor{blue}{This work has been accepted by IEEE TGRS for
  publication in a future issue.}
  Choosing how to encode a real-world problem as a machine learning task
  is an important design decision in machine learning.
  The task of glacier calving front modeling has often been approached as a
  semantic segmentation task.
  Recent studies have shown that combining segmentation with edge detection can
  improve the accuracy of calving front detectors.
  Building on this observation, we completely rephrase the task as a contour tracing
  problem and propose a model for explicit contour detection
  that does not incorporate any dense predictions as intermediate steps.
  The proposed approach, called ``Charting Outlines by Recurrent Adaptation'' (COBRA),
  combines Convolutional Neural Networks (CNNs) for feature extraction and
  active contour models for the delineation.
  By training and evaluating on several large-scale datasets of Greenland's outlet glaciers,
  we show that this approach indeed outperforms
  the aforementioned methods based on segmentation and edge-detection.
  Finally, we demonstrate that explicit contour detection has benefits over pixel-wise methods
  when quantifying the models' prediction uncertainties. 
  
  The project page containing the code and animated model predictions can be found at
  \url{https://khdlr.github.io/COBRA/}.
\end{abstract}

\IEEEpeerreviewmaketitle
\section{Introduction}

\IEEEPARstart{R}{ecent} years have seen
rapid warming in the Polar regions, which has led to an exceptionally
large mass loss of the Greenland Ice Sheet~\cite{mouginot2019_fortysix}.
This loss of ice mass translates into global sea level rise
and can cause feedback effects that further increase warming of the Arctic~\cite{tedesco2016_darkening}.
Closely monitoring the Greenland Ice Sheet is therefore of paramount importance.
About half of this ice mass loss is generally attributed to the glacier dynamics, like dynamic imbalance and increased discharge.
The remaining half are attributed to negative surface mass balance, which mostly stems from an increase in surface melt~\cite{shepherd2020_mass}.

Following the rapid changes in air and sea temperature,
glacier dynamics in these regions are changing quickly.
One essential indicator for dynamic changes of marine-terminating glaciers
is the calving front, which is the boundary line of the glacier
from which ice bergs calve off.
In order to better understand the glaciological processes and
provide more accurate constraints for glacier modelling,
a detailed monitoring of the glaciers' calving fronts is necessary.
With the ever-growing availability of satellite remote sensing data,
monitoring glaciers at a large scale with high temporal frequency 
has become possible, but requires automated methods.
Therefore, recent years have seen rapid advances in
applying machine learning for glacier monitoring,
which will be explored in more detail in section~\ref{sec:related-calving}.

With the rise of deep learning methods in remote sensing,
the predominant method of approaching this task has been via \emph{semantic segmentation}.
In this formulation, each pixel is assigned a label that corresponds to either the
\emph{glacier} class or the \emph{sea} class.
Given the large number of studies on semantic segmentation in computer vision,
the methods and models for this task are well understood and provide decent results
when applied to calving front detection.
However, these methods require post-processing steps to extract
the actual calving front from the segmentation masks.

\begin{figure}
\begin{center}
  \includegraphics[width=\linewidth]{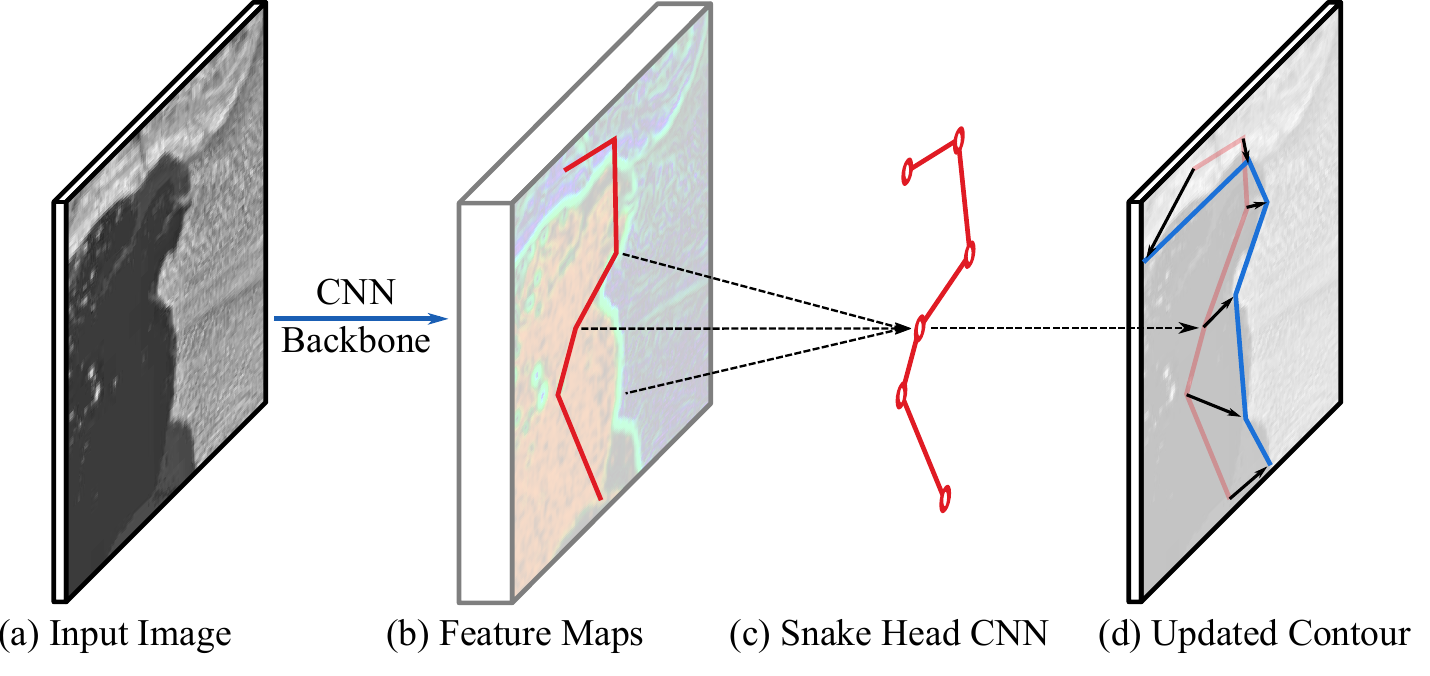}
\end{center}
\caption{High-level overview of our deep active contour model for delineating calving fronts.
  First, the backbone network takes the input image (a) and derives feature maps.
  Then, a sample is taken from these feature maps at the position of each vertex (b).
  These features are evaluated by the Snake Head (c) which predicts offsets for each vertex.
  Finally, the offsets are applied to update the contour (d).
  This process is repeated multiple times.
}
\label{fig:teaser}
\end{figure}

Noting that segmentation is only a proxy for the actual task of calving front detection,
and neither the sea or the inward glacier area are of actual interest for calving front detection,
the field has seen a recent trend towards edge detection methods.
By combining computer vision methods for pixel-wise edge detection
with the aforementioned segmentation task,
predictions are thus greatly improved~\cite{cheng2021_calving,heidler2022_hedunet}.

The goal of this study is to provide a new angle on this task.
Picking up the trend towards edge detection,
we propose to completely move away from pixel-wise prediction architectures
and rethink the task from the ground up.
The desired final prediction format for calving front detection is a vectorized polyline,
which is a data format that is well suited for downstream analysis and modeling applications.
Therefore, we are looking to build a model that directly outputs the calving fronts in
this desired format instead of recovering the vectorized contour from intermediate predictions.
By radically redesigning the neural network architecture,
we are able to move away from pixel-wise classifiers
and instead arrive at a model that directly predicts the calving front as a polyline.

This approach has several theoretical benefits over representing the desired output
by a dense, pixel-wise mask.
(1) As the predictions are already in a vector format,
there is no need for complicated post-processing pipelines like with pixel-wise approaches.
(2) By its very design, the model will learn to focus on the actual object of interest,
the calving front.
(3) Looking closer into the application, explicit contour prediction
provides a natural way of encoding prior knowledge into the network.
In pixel-wise detection frameworks,
the network may predict undesired outputs like disconnected line segments.
By directly predicting an explicit contour, such issues are eliminated.
(4) Explicit contours make more efficient use of computational resources.
A sequence of vertices takes fewer parameters than a dense mask.
(5) Finally, the vectorized representation allows for a better quantification of model uncertainty
as the joint probability distribution of a sequence of vertices is easier to model
than that of pixel-wise masks.

Convinced by these theoretical considerations,
we set out to developing a calving front detection model
that directly predicts the desired contours.
Contour-based approaches for the segmentation of regions in natural images
have been extensively studied in the form of \emph{Active Contours},
which are also called \emph{Snakes}~\cite{kass1988_snakes}.

In order to provide robust and stable calving front predictions for
downstream applications such as glaciological studies and models,
it is important to quantify the reliability of the model's predictions.
Therefore, we also explore the question of uncertainty quantification
in calving front detection.
Experimental results using the the Monte Carlo Dropout
method~\cite{gal2015_dropout} across different models
suggest that uncertainty quantification with contour-based models
can indeed bring benefits over pixel-wise models.

Overall, we summarize the goals and contributions of this study by the following points:
\begin{enumerate}
  \item We rephrase the task of automated glacier calving front detection
    from a segmentation task to a contour detection task
    and show that deep active contour models are a feasible approach to solving this task.
  \item We develop a specialized deep active contour model for 
    the delineation of glacier calving fronts
    which outperforms both pixel-based approaches
    as well as existing deep active contour models.
    The effect of the design decisions is validated through extensive ablation studies.
  \item We explore the benefits of contour-based methods for
    uncertainty quantification compared to pixel-wise methods.
\end{enumerate}

\section{Related Work}\label{sec:related}
In order to place our work into the context of existing research,
we provide a brief overview of existing calving front detectors,
as well as methods for explicit edge predictions.

\subsection{Detecting Calving Fronts in the Deep Learning Era}\label{sec:related-calving}
Given the strong performance of deep learning-based methods
for calving front detection,
traditional vision methods have largely become insignificant for
this task~\cite{baumhoer2018_remote,heidler2022_hedunet}.
Therefore, this section focuses on the deep learning-based methods.
Here, most approaches formulate the task as a
variant of sea-land segmentation.
In this formulation, a semantic segmentation network is used to separate the image into
land and ocean classes~\cite{baumhoer2018_remote}.
There is a considerable number of well-tested segmentation architectures like UNet~\cite{ronneberger2015_unet},
which is a strong baseline for most segmentation tasks.
Even without any changes to the network itself,
this approach can yield satisfactory results for calving front detection,
which has been shown in previous studies for both
the Greenland Ice Sheet~\cite{mohajerani2019_detection} and
the Antarctic Ice Sheet~\cite{baumhoer2019_automated}.
Seeing this strong baseline performance of the UNet,
Periyasamy et al.~\cite{periyasamy2022_how} show that the performance of such a model
can be greatly improved by tweaking network components like
normalization layers, the loss function, or dropout rate.

Further progress in this field was made by extending the UNet model
or exploiting the advances of more recent segmentation model architectures.
For example, Loebel et al.~\cite{loebel_extracting} add more layers
and thereby increase the number of down- and upsampling steps.
This enlarges the spatial context that the network can consider for its decisions
and therefore leads to better predictions.
Following recent advances in Transformer-based model architectures,
Holzmann et al.~\cite{holzmann2021_glacier}
enhance the UNet model with attention gates to improve
the interpretability of the model and better understand the learning process.
Another newer neural network architecture that has successfully been adopted for
calving front detection is DeepLabv3+~\cite{chen2018_encoderdecoder}.
Both Zhang et al.~\cite{zhang2021_automated}, Cheng et al.~\cite{cheng2021_calving}
and Gourmelon et al.~\cite{gourmelon_2022}
bring in ideas from this architecture
to obtain more accurate delineations of glacier calving fronts.
Finally, it is also possible to combine image classification and segmentation~\cite{marochov2021_image},
which can lead to more robust results.

Recently, there appears to be a trend towards models that
approach calving front detection by extending or even replacing semantic segmentation
with edge detection methods.
By focusing on the boundary between the two classes rather than the areas of sea and glacier,
these models are encouraged to learn features that are informative of the calving front
rather than features of the sea and glacier areas.

Wavelet transforms are one possible approach which makes use of the abrupt changes in
texture between glacier and sea to determine the location of the calving front~\cite{liu2021_automated}.
Davari et al.~\cite{davari2022_pixelwise} use a different transformation,
namely the Euclidean Distance Transformation,
and train a network that predicts each pixel's distance to the calving front
instead of a binary class.
Great potential lies especially in the combination of segmentation and edge detection.
Both HED-UNet~\cite{heidler2022_hedunet} and the Calving Front Machine (CALFIN)~\cite{cheng2021_calving}
choose this approach to outperform models that focus on only one of these two aspects.

Contrary to these models, our approach is not to predict pixel-wise segmentation or edge masks,
but instead to explicitly predict a contour parameterized by a fixed number of vertices.

\subsection{Explicit Edge Prediction}
The idea of explicitly parameterizing contours in an image was pioneered quite early in the history
of computer vision by Kass et al.~\cite{kass1988_snakes}.
They proposed \emph{Active Contours} or \emph{Snakes},
which evolve from an initial contour by iteratively minimizing an energy functional.
By design, this functional takes its minimum when the contour coincides
with the desired boundary in the image.
Using Radarsat data, this approach has been shown to work for
the delineation of Antarctic coastline on a coarse scale~\cite{klinger2011_antarctic}.
The main drawback of conventional active contour methods is the fact that
they are limited to single-channel imagery without any natural extension to multi-channel imagery.
Further, they are sensitive to local image contrast and
the results depend highly on the initialization of the contour.

As automatic feature extraction is a strong suit of deep learning models,
the idea of combining active contours with deep learning is not a new one.
Rupprecht et al.~\cite{rupprecht2016_deep} introduced a deep active contour model that
works by first predicting a two-dimensional offset field that points from each pixel
toward the closest boundary point.
An initial contour will then evolve along this offset field until it converges.
However, this method is not end-to-end trainable as it relies on the intermediate offset field
and no gradients flow through the actual curve evolution.
While this approach can work on multi-channel imagery,
it still suffers from a strong dependence on contour initialization.

In an effort to introduce an end-to-end trainable deep active contour model,
Peng et al.~\cite{peng2020_deep} proposed to make not only the feature extraction part learnable,
but the contour evolution step as well.
Their model, termed \emph{Deep Snake} first derives feature maps
using a convolutional backbone network and then
samples the features at the position of each vertex.
From these sampled features, a one-dimensional CNN then predicts the offsets for each vertex.
Like with conventional active contour models,
this process is then iterated to refine the predictions.

As one of the most recent models in this line of research,
Deep Attentive Contours (\emph{DANCE})~\cite{liu2021_dance} improves on the Deep Snake idea
by introducing an ``edge attention map'', which influences the speed of the snake evolution.
While vertices far from the target boundary are evolving quickly,
the update speed for points closer to the boundary is slowed down.

Inspired by these advances, our goal is to develop an
active contour model for the task of calving front detection.
The existing models address the computer vision task of instance segmentation,
where objects in an image are locally segmented.
For calving front delineation however, one global line between glacier and ocean is needed.
This disparity and further differences, like the general shapes of the objects of interest,
call for a completely different network architecture as well as changes to loss functions and to
the network training protocol.

\section{Deep Active Contour Models for Calving Front Delineation}
When approaching the task of calving front detection,
we first take a look at how a human would proceed in solving the task.
In discussions with experts and when annotating calving fronts ourselves,
one central observation is the order in which different areas in the scene are addressed.

To a human annotator,
it does not make much difference whether they are told to
trace the edge between two objects in an image or
fill in the areas that both objects occupy.
In both cases, they will usually start out by tracing
the boundary area between the two classes with minute attention to detail.
When asked to perform segmentation,
the remaining areas are then filled in with broad strokes in a second step.
So while humans will approach both tasks in a similar fashion,
a focus on the boundary appears to be the more natural way
to formulate the task of calving front detection.

For a neural network model,
the way a task is formulated changes everything.
As these models are trained to minimize some loss function,
the final performance is determined by how well
a low loss value translates to good performance on the actual task.
For instance, it has been observed that the cross-entropy loss
used for training semantic segmentation models can lead to
blurry edges between the classes due to the fact that each pixel
contributes equally to the final loss,
no matter its position in relation to the objects in the image.
This implies that a model can minimize most of its loss by
correctly classifying the simpler pixels that lie in the interior of
the objects of interest.
In turn, the model will spend less attention on the pixels near the boundary, which are
much more important for solving the task~\cite{drozdzal2016_importance,heidler2022_hedunet}.

Phenomena like these are likely the reason that calving front delineation
has recently seen a shift towards improving these segmentation models
by including pixel-wise edge detection tasks.
By putting more focus on the edges, the model is forced to learn how to better distinguish
the classes in these critical areas~\cite{cheng2021_calving,heidler2022_hedunet}.
Instead of combining segmentation with edge detection in a pixel-wise framework,
we take a more radical approach in this study.
By completely eliminating the semantic segmentation aspect and focusing only on the edges,
it is possible to reformulate the task in such a way
that it does not require pixel-wise classifications.
Instead the model will directly output a vectorized contour.

\begin{figure}
\begin{center}
  \includegraphics[width=\linewidth]{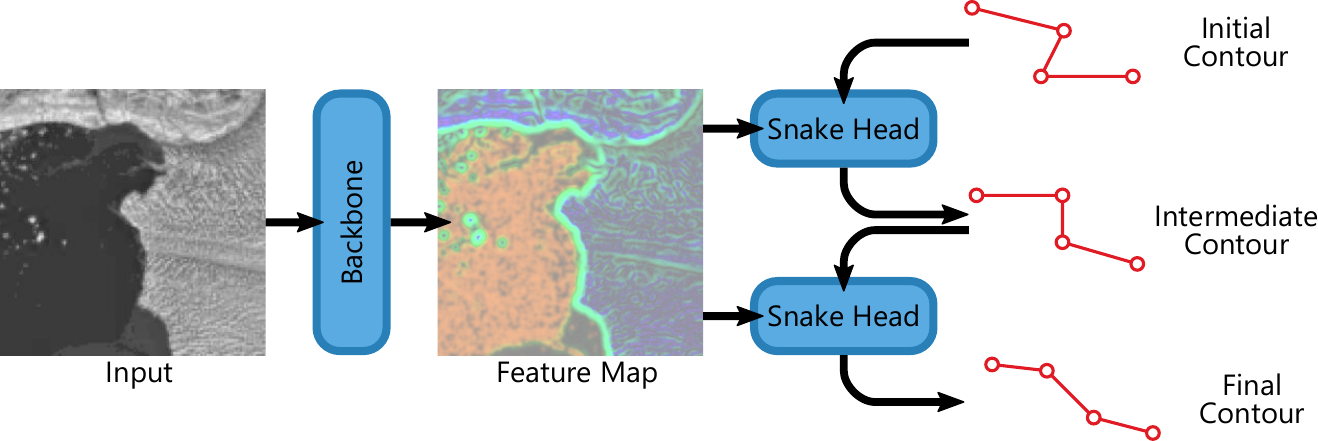}
\end{center}
\caption{%
  Architecture overview of our model.
  Note that while this diagram shows only two iterations of the Snake Head,
  the number of iterations is actually an arbitrary hyperparameter,
  which we set to 4 for our experiments.
}
\label{fig:architecture}
\end{figure}

\subsection{General Model Architecture}
Inspired by the ideas of Peng et al.~\cite{peng2020_deep},
we develop a deep contour model for the delineation of glacier calving fronts.
As is common in many recent computer vision models,
our model consists of two main components which perform different subtasks
in order to solve the overall task together.
The \emph{backbone} is a general-purpose two-dimensional CNN which is used to
extract semantically valuable features from the input imagery.
The second component is a \emph{prediction head}, which makes use of the
backbone's features to derive the final network predictions.
In our model, the prediction head takes the role of the active contour iteration.
Therefore, it will take a contour and the backbone's feature maps as its inputs,
and update the contour to better match the desired boundary.
Due to this functional similarity, we call this component the \emph{Snake Head}.
The overall architecture of the network is visualized in Figure \ref{fig:architecture}.
Notably, this framework can be trained end-to-end, as all components are fully differentiable.

For the backbone, multiple feature extractors were evaluated.
Initial experiments with standard ResNet backbones produced unsatisfactory results.
This leads us to believe that while ResNets are a strong backbone for many vision tasks,
they are likely not optimal for deep active contour models.
In search of a better suited backbone network,
the Xception backbone~\cite{chollet2017_xception}
used by Cheng et al.~\cite{cheng2021_calving}
in their study of Greenland's glaciers
proved to be a very capable backbone for remote sensing of glaciers
that transfers well to deep active contour models.

\subsection{Snake Head}
\begin{figure}
\begin{center}
  \includegraphics[width=\linewidth]{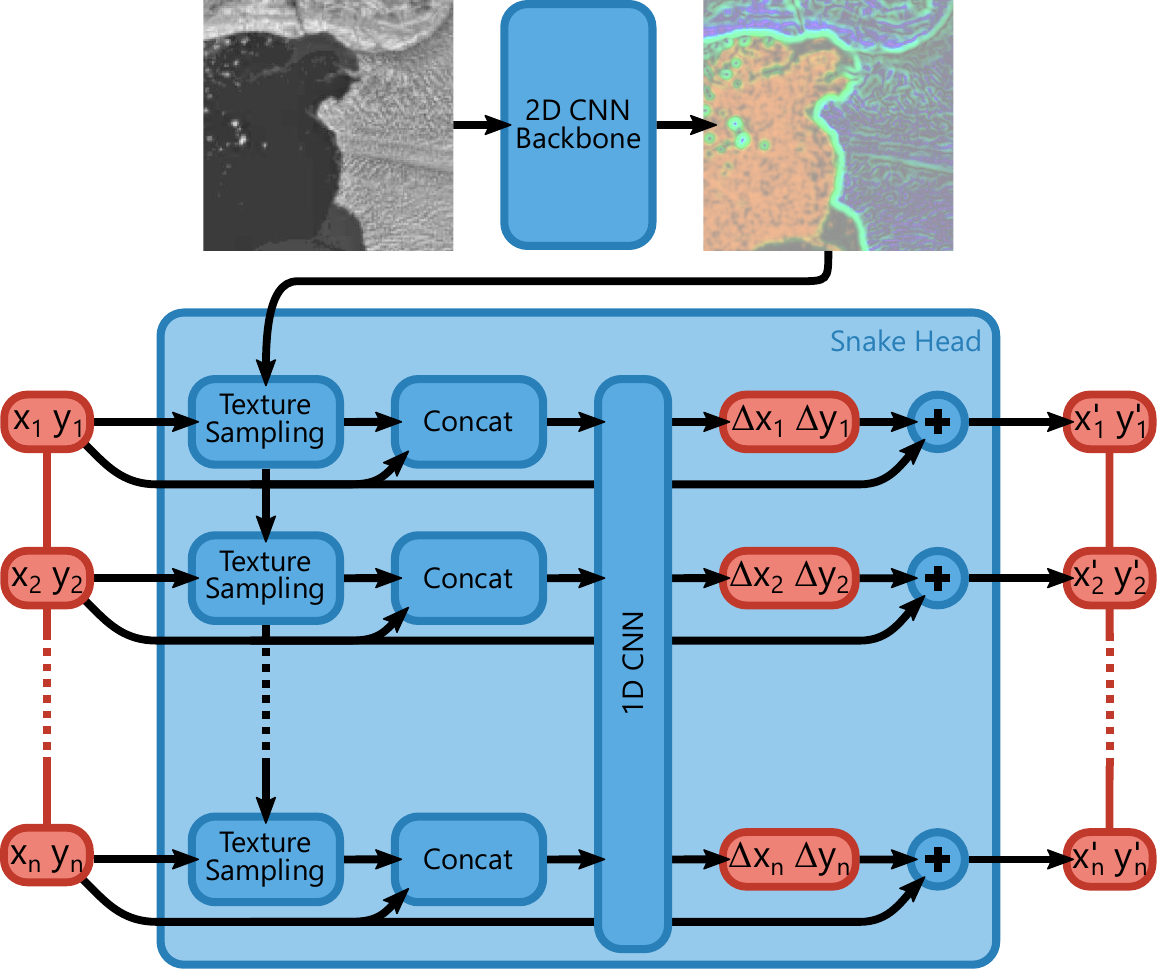}
\end{center}
\caption{%
  Detailed view of the Snake Head.
  After the feature maps are sampled at the vertex positions,
  the vertex coordinates are concatenated to the vertex features.
  The 1D CNN then predicts offsets for each vertex.
  These offsets are added to the input coordinates to obtain the Snake Head's output.
}
\label{fig:snakehead}
\end{figure}

The central challenge in predicting contours from an image is the fact that
input and output are represented in different dimensionalities.
While the input image is represented by a two-dimensional grid of pixels,
the contour that the model should output is
given as a sequence of vertices,
which is one-dimensional.
The idea of active contour models is to start with an initial contour and then
iteratively update this contour based on the image values at each vertex.
Conceptually, deep active contour models do nearly the same thing.
However, they do not directly sample the image values,
but instead sample the values from the feature map derived by the backbone network.

After sampling the backbone features at the vertex positions,
the Snake Head predicts an offset for each vertex,
which represents how the vertex needs to be shifted so that
the entire sequence of vertices can better represent the
true contour.
This is achieved by using a one-dimensional CNN.
While conventional, two-dimensional CNNs pass information between
adjacent pixels,
the one-dimensional CNN used in the Snake Head passes information
between adjacent vertices of the contour.
In order to enable the passing of information between vertices that
are far away from each other,
we stack multiple such convolutional layers.
The receptive field of the snake head is further increased
by using dilated convolutions.
In our model, the Snake Head is therefore given as a stack of
dilated convolutions. 
We set the sequence of dilation rates to 1, 3, 9, 9, 3 and 1,
as similar setups have proven to be successful at capturing
low- and high-frequency features in signal processing tasks~\cite{vandenoord2018_pwavenet}.

In order for the Snake Head to gain some spatial reasoning capabilities,
we also include the vertex coordinates as additional input features for the
Snake Head CNN.
This allows the model to not only ensure a homogeneous spacing of the output vertices,
but also to learn some prior assumptions on the shape of the calving fronts.
The overall working mechanism of the Snake Head is shown in Fig.~\ref{fig:snakehead}.

To translate the iterative nature of the active contour method,
we apply the Snake Head multiple times with the shared weights
to obtain more refined predictions.
Starting with the initial contour,
the Snake Head samples features at the vertex positions,
calculates and applies the offsets,
and repeats the process.
Compared to conventional active contour models,
which can take dozens~\cite{klinger2011_antarctic}
or even hundreds~\cite{wang2017_active} of iterations to converge,
the deep active contour model converges to satisfactory results after
a small number of iterations.
For our experiments, we set the number of iterations to 4.

In the context of deep neural networks,
the Snake Head can also be regarded as a recurrent neural network.
The locations of the vertices then represent the hidden state of the
network, which is updated throughout the iteration steps until
the final output is achieved.

\subsection{Loss Function}
The loss function is a crucial element of any deep learning model,
as it measures how well the model is performing and gives feedback for improving
the network via backpropagation.
When predicting contours, the loss function should therefore measure the
similarity between the predicted contour $p$,
represented by vertices $p_i$ with $1 \leq i \leq V$,
and the true contour $t$, given by the vertices $t_j$ with $1 \leq j \leq V$. 

Common loss functions for polygon regression are based on the $L_1$ and $L_2$ losses,
which, following the above notation, are defined as follows:
\begin{align}
    L_1(p, t) &= \frac{1}{V} \sum_{i} \|p_i - t_i\|\\
    L_2(p, t) &= \frac{1}{V} \sum_{i} \|p_i - t_i\|^2
\end{align}

These loss functions have a fundamental issue when predicting glacier front lines.
By only computing the distance between the vertices of $p$ and $t$ at the same index,
they tacitly assume that each predicted vertex corresponds to
exactly one vertex in the true contour.
However, the model has no way of knowing how the
ground truth vertices were placed along the true contour.
In the setting of Deep Snake and DANCE, the vertices were placed equidistantly
along the contour of objects that were largely convex,
so this assumption did not have much negative impact there.

When predicting glacier frontlines however,
this issue becomes much more prominent due to
the irregular and jagged shape of these contours.
As the model essentially tries to minimize the $L_1$ or $L_2$ loss
for a number of possible parameterizations of the true contour at the same time,
the resulting predictions lack sharp edges and instead follow a
smoothed version of the true outline.

Naturally, contour prediction is not the first task to
face challenges like these.
For example in the context of time-series analysis,
slight variations in timing are often less important
than the general shape of the time-series.
\emph{Dynamic time warping} (DTW) is a method that was proposed
by Sakoe and Chiba~\cite{sakoe1978_dynamic}
in order to address this very issue.
Given two sequences, they not only compare the pairwise differences,
but instead first find an optimal alignment between the two sequences
and then calculate the distances based on that alignment.

In our setting, the parameterization of a contour takes the
role of time in the original DTW.
Formally, we define the DTW loss for two contours $p$ and $t$ to be

\begin{equation}\label{eq:dtw}
  \mathop{\mathcal{L}_{\text{DTW}}}(p, t) =
  \min_{(i_k, j_k)_{k\in[K]} \in \mathcal{K}} \sum_k \|p_{i_k} - t_{j_k}\|_2^2\,,
\end{equation}
where $\mathcal{K}$ denotes the set of all possible re-alignments $(i_k, j_k)_{k\in[K]}$
that satisfy the following three conditions:
\begin{itemize}
  \item For any $i \in \{\,1, \dots, V\,\}$ there is a $k$ with $i_k = i$.
  \item For any $j \in \{\,1, \dots, V\,\}$ there is a $k$ with $j_k = j$.
  \item The sequences $i_k$ and $j_k$ are non-decreasing in $k$.
\end{itemize}
Under these conditions, the DTW loss can be efficiently calculated
using dynamic programming~\cite{sakoe1978_dynamic}.

A possible issue with the use of DTW as a loss function
in deep learning is the fact that it is not smooth
due to the minimum operator applied in Eq.~\ref{eq:dtw}.
Seeing this, Cuturi and Blondel~\cite{cuturi2018_softdtw}
replace the minimum with a \emph{soft minimum} which they define as
\begin{equation}
 \operatorname{softmin}_\gamma(x_1, \dots, x_n) = -\gamma \log \sum_{k=1}^n \exp{\frac{-x_k}{\gamma}},
\end{equation}
with a smoothness parameter $\gamma > 0$.
In the limit $\gamma \to 0$, the conventional minimum operator is recovered.

\subsection{Implementation Details}
A central issue with naively backpropagating
the loss through the snake iteration is the fact
that the early iterations show poor convergence to the target contour.
n This is easily fixed by stopping the gradient
from flowing through the
coordinates at the beginning of each snake step.
To still encourage quick convergence
of the contours during inference,
we leverage deep supervision by
including an additional loss term for each intermediate step. 
During training,
the current contour is compared to the ground truth after each
snake step,
and the resulting loss is added towards the final loss for
the gradient calculation.
Unless otherwise stated, all models use a contour parametrization by 64 vertices.

All models in the study were trained for 500 epochs on the training dataset.
We used the Adam optimizer~\cite{kingma2015_adam} with an initial learning rate of $10^{-3}$
decaying to $4 \cdot 10^{-5}$ on a cosine decay schedule~\cite{loshchilov2017_sgdr}.

Our models are implemented in JAX~\cite{jax2018github} using the
Haiku framework~\cite{haiku2020github}.
Training was conducted on a single NVIDIA RTX 3090 GPU
with 24GB of VRAM.
Training an instance of the model,
took around 25 hours,
and had an estimated energy consumption of \SI{8.1}{\kWh}.

\section{Experiments \& Results}
\begin{figure*}
\begin{center}
  \includegraphics[width=\linewidth]{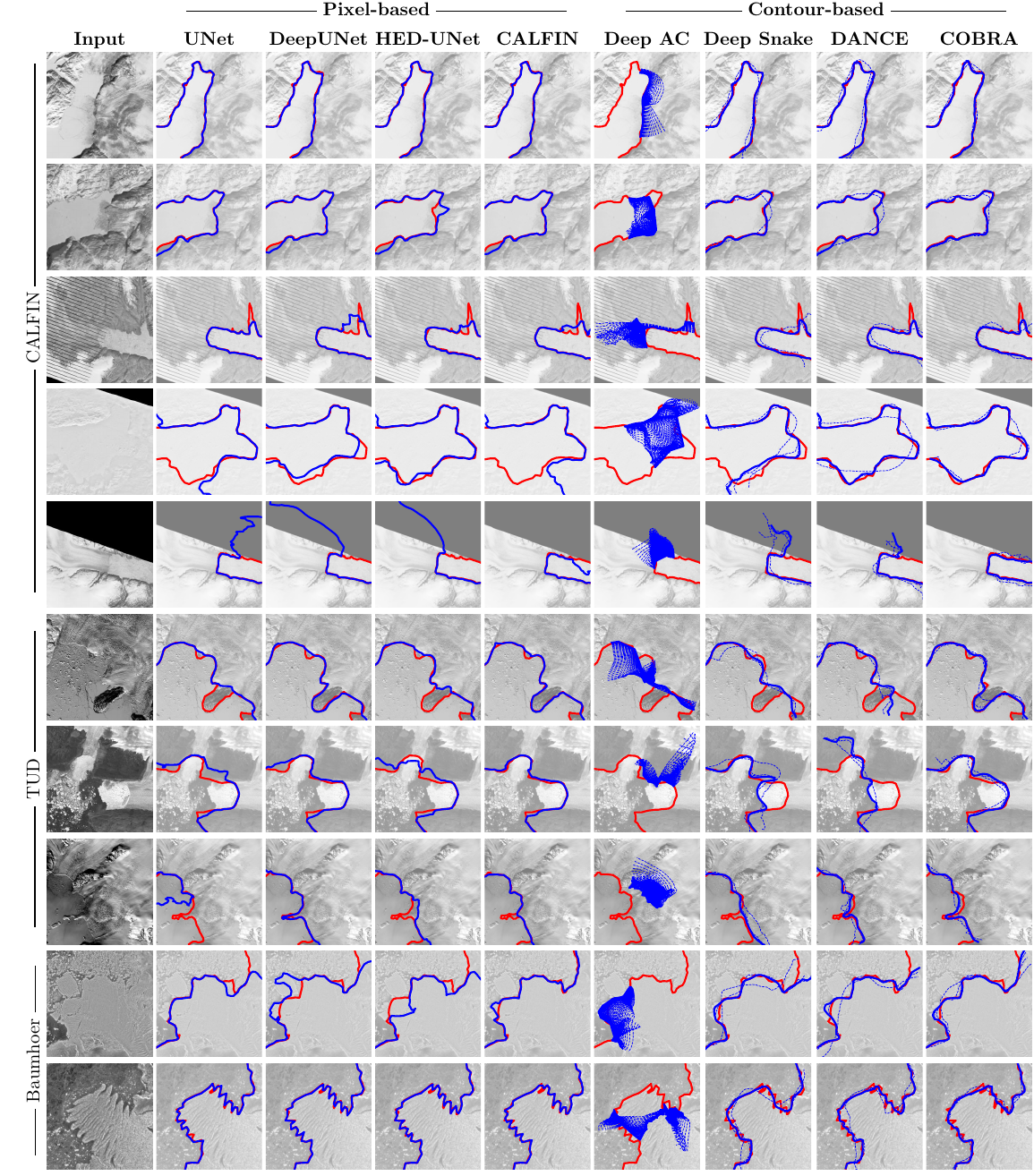}
\end{center}
\caption{%
  Visualization of prediction results (blue) for the different models
  and corresponding ground truth (red) on the test datasets.
  For the iterative, contour-based models intermediate results are 
  displayed as dashed blue lines.
  Best viewed in color.
}\label{fig:results}
\end{figure*}

\subsection{Datasets}
In order to thoroughly evaluate our model and compare it with other approaches,
we choose two large-scale datasets of marine-terminating glaciers in Greenland
for training and evaluation purposes,
namely the \emph{CALFIN} dataset~\cite{cheng2021_calving}
and the calving front dataset from \emph{TU Dresden} (TUD)~\cite{loebel_extracting}.
Both of these datasets include respective testing data.
Further, the Baumhoer dataset~\cite{baumhoer2019_automated}
consists of synthetic aperture radar (SAR) data of Antarctic glaciers,
thus serving as a benchmark for the models' ability to generalize to a different
data modality and a different ice sheet.

\subsubsection{CALFIN Dataset}
The CALFIN dataset consists of near-infrared data from the various Landsat missions,
and is most notable for the long time span of acquisition times,
ranging from 1972 to 2019.
Its spatial coverage is 66 Greenlandic glaciers,
which amount to 1541 Landsat scenes.
In an effort to improve generalizability to different sensors,
the training dataset also includes 232 single-polarization 
Sentinel-1 scenes of glaciers in Antarctica.
The corresponding test dataset consists of 162 Landsat near-infrared scenes. 
For all of the mentioned scenes, the calving fronts were manually delineated.

\subsubsection{TU Dresden Dataset}
In contrast to this, the TUD dataset puts its focus
on the eighth iteration of the Landsat mission,
providing a dense time-series of recent acquisitions of
Greenland's marine terminating glaciers.
The captured scenes range from 2013 to 2021 for a total of 1127 tiles.
For studies related to feature importance and data fusion,
it includes the full multispectral imagery, as well as topography data
and texture information derived using gray-level co-occurrence matrix statistics.
For interoperability with the other datasets,
only the panchromatic imagery is used in this study.

\subsubsection{Baumhoer Dataset}
Another test set for evaluating the generalization of the trained models is given by the
Baumhoer Dataset~\cite{baumhoer2019_automated}.
This dataset is vastly different from the other datasets,
as the imagery is not from Greenland, but from Antarctica instead.
Further, it consists of Sentinel-1 SAR imagery, which marks a second challenge in generalization.
While the original dataset is not openly available, an evaluation subset is distributed
along with the CALFIN dataset~\cite{cheng2021_calving}. In order to keep this study fully reproducibile, we only use this publicly available subset of the Baumhoer dataset.
The used testing set consists of 62 Sentinel-1 scenes of glaciers in Antarctica
from the year 2018.

\subsection{Evaluation Metric}
As there is no uniquely defined distance metric between two curves,
many different metrics are being used for evaluating the accuracy of
predicted glacier frontline positions~\cite{baumhoer2018_remote}.
In our work, we adapt the Polis metric~\cite{avbelj2015_metric},
which was originally proposed for measuring the dissimilarity between building footprints.
For two polylines $v$, $w$, with $I$ and $J$ vertices, respectively,
it is defined as the average distance of any vertex
to the respective other polyline.
\[
  p(v, w) = \frac{1}{I} \sum_{i=1}^I d(v_i, w) + \frac{1}{J} \sum_{j=1}^J d(w_j, v),
\]
where $d(v_i, w)$ denotes the distance between vertex $v_i$
and the closest point on the polyline $w$.
Note that this closest point $w$ does not need to be a vertex,
but may be a point between vertices as well.

Compared to other existing metrics like the Fréchet distance~\cite{alt1995_computing},
which is defined as the solution to a min-max problem,
the Polis metric is more easily interpretable
as the ``average'' distance between the two curves.
Further, it was chosen due to its symmetry
and the fact that it takes all predicted points into consideration.

\subsection{Comparison with other models}
\begin{table}
  \centering
  \caption{Mean Deviation (Polis Metric)
  of the Trained Models on the Evaluated Test Sets.}\label{tab:comparison}
  \vspace{-12pt}
  \include{results}
\end{table}
For our comparison study, 
we train a number of different models in order to compare their performance on the test datasets.
To compare with the state of the art in calving front detection and contour-based outline detection,
we include both pixel-wise and contour-based models.

\subsubsection{Pixel-wise Models}
This first group of methods consists of pixel-wise segmentation models
that are known to work well for calving front detection.
\paragraph{UNet~\cite{ronneberger2015_unet}}
This model is a popular semantic segmentation model
that serves as a strong baseline for many segmentation tasks.
It has been successfully applied to calving front detection~\cite{baumhoer2019_automated}.
\paragraph{DeepUNet~\cite{loebel_extracting}}
The model developed and used by the authors of the TUD dataset.
Its main difference from the original UNet model is the addition of two down- and upsampling steps,
which make the model deeper and more aware of spatial context.
\paragraph{HED-UNet~\cite{heidler2022_hedunet}}
A combination of the UNet model with an edge detection model, HED-UNet was
originally developed to detect glacier frontlines
on the Baumhoer dataset~\cite{baumhoer2019_automated}.
\paragraph{Calving Front Machine (CALFIN)~\cite{cheng2021_calving}}
This model was introduced by the authors of the CALFIN dataset.
The model is based on the segmentation architecture DeepLabv3+~\cite{chen2018_encoderdecoder}.
It is the first to leverage the potential of the Xception network for calving front detection.
\subsubsection{Contour-based Models}
For a comparison to existing contour-based models,
we also include models from this group into the comparison.
It should be noted that unlike the pixel-wise models above,
they were not developed for calving front detection.
\paragraph{Deep AC~\cite{rupprecht2016_deep}}
One of the first works to combine active contour models with deep learning,
this model uses a two-dimensional CNN to predict an offset field that points towards
the nearest contour point from each pixel and then evolves a contour along this offset field.
\paragraph{Deep Snake~\cite{peng2020_deep}}
Originally proposed as a contour-based model for instance segmentation,
we have made slight changes to this model to perform calving front detection.
Specifically, the circular convolutions in the network were replaced
with regular 1D convolutions, as the predicted contours for calving fronts
should be open polylines and not closed polygons.
Further, the object detection head of the model was removed
as with calving fronts, there is always exactly one contour to be predicted.
For a fair comparison, we train and evaluate this model not only with the ResNet-50 backbone,
but also with the Xception backbone.
\paragraph{Deep Attentive Contours (DANCE)~\cite{liu2021_dance}}An iteration of the Deep Snake~\cite{peng2020_deep} model,
DANCE introduces an edge attention map that
speeds up the evolution for vertices far from the true edge and
slows the evolution for vertices on the true edge.
We applied the same adaptations to this model as to the Deep Snake model.

As CALFIN provides the largest and
longest record of glacier observation,
we train all models on the CALFIN training set,
and then evaluate them on the CALFIN,
TUD and Baumhoer test sets.
The numerical evaluation results for
this comparison study are displayed in Table~\ref{tab:comparison},
and some visual results are displayed in Figure~\ref{fig:results}.
For the proposed COBRA model,
we train 3 randomly initialized models
and report mean and standard deviation across these 3 runs.

Comparing the pixel-wise models,
we can reproduce the increased performance of the calving front-specific models
over the baseline UNet.
All three of these models, namely DeepUNet, HED-UNet and the Calving Front Machine
cut down the average prediction error from UNet's \SI{224}{\metre}
to the range of \SIrange{130}{138}{\metre} on the CALFIN test set.
There is however a difference in generalization to the other datasets,
where the DeepUNet seems to generalize best to SAR imagery,
and the Calving Front Machine generalizing better than others on the TUD dataset,
which is also based on Landsat imagery.

Looking at the contour-based models, the Deep AC model falls behind the competition,
and performs the worst out of all the models in our experiments.
The bad performance of the Deep AC model is easily explained when looking at
the visual results in Figure~\ref{fig:results}.
It stems from the fact that its predictions tend to only represent a part of the desired
curve, as there is no regularization term that forces the prediction to cover the entire
calving front.

With the exception of the Deep AC Model~\cite{rupprecht2016_deep},
the contour-based approaches perform considerably better on the CALFIN evaluation
than their pixel-based counterparts,
especially when using the Xception backbone network.

On the other hand,
generalizing to the Antarctic Baumhoer dataset is particularly hard
for the contour-based methods.
We attribute this to the presence of
jagged floating ice-tongues (cf.~Figure~\ref{fig:results}, last row),
which are not observed in the same way during training.
Such features lead to more complex outlines,
which need more vertices for their representation,
as can be seen from the results of the study on vertex numbers
in section \ref{sec:vertices}

Overall, our proposed network outperforms
both the pixel-based and
other contour-based models
by a considerable margin
on the CALFIN and TUD evaluations.
Even when generalizing to the radically different
Baumhoer dataset, our model still maintains a respectable performance.

Inference results from the COBRA model for the three testing datasets are available for
online viewing and as a shapefile download at 
\url{https://github.com/khdlr/COBRA/tree/master/inference_results}.

\subsection{Quantifying Uncertainty with Contour Models}
\begin{figure*}
  \begin{center}
    \includegraphics[width=\linewidth]{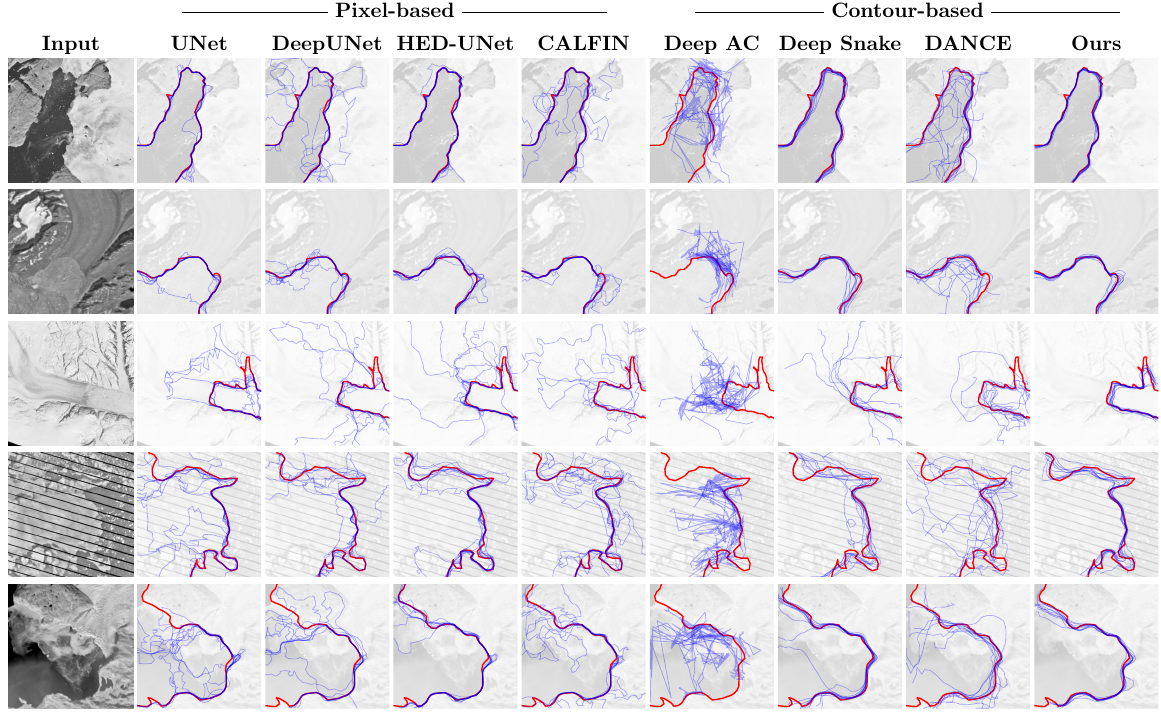}
  \end{center}
  \caption{
    Visualization of posterior samples obtained using Monte Carlo dropout (blue)
    from the different models overlaid on top of the ground truth (red) for scenes from the CALFIN test set.
  }
  \label{fig:uncertainty}
\end{figure*}

With deep learning models growing ever more complex,
quantifying the uncertainty of their predictions
at inference time
has become an important consideration
when working with such models.
Deep learning models being over-confident in their predictions
is a common issue~\cite{guo2017_calibration}.
As the models are usually trained on definite ground truth,
the models are never taught to concede their uncertainty in ambiguous cases.

One elegant method for the quantification of network uncertainties is
known as Monte Carlo (MC) Dropout.
In their seminal study, Gal and Ghahramani~\cite{gal2015_dropout} demonstrated
that a deep learning model trained with dropout layers
can be interpreted as approximated Bayesian inference in a deep Gaussian process.
Samples from the posterior distribution approximated by such a model can be recovered quite easily by
enabling the dropout layers not only at training time,
but also at inference time.
It has been shown that MC dropout can quantify model uncertainties
well for remote sensing tasks like
aerial image segmentation~\cite{dechesne2021_bayesian}.
Recently, Hartmann et al.~\cite{hartmann2021_bayesian}
also successfully applied a Bayesian UNet for the segmentation of glaciers
in SAR imagery.

As the MC Dropout (MCD) method is simple to implement and evaluate
compared to other uncertainty quantification methods,
we choose this approach for quantifying uncertainties in the model predictions.
In order to estimate the hardness of samples at inference time
and get an estimate for the model uncertainty,
we calculate the original, deterministic model prediction, as well
as multiple additional predictions using the MCD technique.
If these predictions all line up well with the original model prediction,
the model can be assumed to be quite certain of its prediction.
On the other hand,
a large deviation between the original model prediction
and the MCD predictions corresponds to ambiguity in the
model output, implying a potentially higher prediction error.

Taking the 10 MC samples and the model's deterministic prediction,
we estimate the model uncertainty
as the average Polis-distance of each MC sample
from the deterministic prediction.

Our hypothesis is that the explicit edge parameterization by vertices
lends itself much better to uncertainty quantification from these posterior samples
than dense pixel-wise predictions,
due to the fact that the explicit representation requires much less parameters.
With less parameters,
the covariance between the parameters becomes more tractable,
and therefore easier to approximate for any model.

For our uncertainty quantification study,
we apply MC dropout with a dropout rate
of 20\% to the aforementioned models.
After training the models,
we draw 10 predictions with enabled dropout (posterior samples)
per model for each test scene
in order to assess the quality of the uncertainty quantification.

Figure~\ref{fig:uncertainty} shows the posterior samples obtained using
the MC dropout models.
It can be observed that the pixel-wise calving front detectors
can collapse completely on hard scenes.
All the evaluated pixel-wise methods suffer from this phenomenon,
suggesting that it is indeed related to the mode of representation.
Inspection of the underlying segmentation masks suggests that this is due
to the fact that when working with segmentation masks,
small changes in the segmentation can completely change the topology of the prediction
as previously connected regions can become disconnected and vice-versa.

Due to this effect, 
estimating the model uncertainty using
MC dropout can overestimate the hardness of the samples
for these models on easy scenes.

\begin{table}
  \center
  \caption{Uncertainty Quantification:
    Pearson Correlation between
    model uncertainty and actual prediction error
    (Polis Metric).
  }\label{tab:pearson}
  \input{pearson.tex}
\end{table}

By their design, contour-based methods do not have this limitation,
as they predict the frontline directly.
Among these models, the DANCE architecture seems affected by similar issues
as the pixel-wise models,
which we attribute to the fact that DANCE incorporates an intermediate dense prediction.
Overall, both Deep Snake~\cite{peng2020_deep} and our proposed model
appear to be best suited for uncertainty quantification using MC dropout,
with very similar samples on easy scenes,
and deviating samples in areas that are harder to delineate.
Among the pixel-based methods, HED-UNet~\cite{heidler2022_hedunet}
appears to be the best at quantifying its uncertainty.

In an effort to numerically evaluate the uncertainty quantification,
we then calculate the Pearson correlation coefficient between
the model uncertainty and the actual prediction error.
A high correlation between these two variables corresponds to better
uncertainty quantification,
as the model should only be certain when its prediction is actually correct,
while a high model uncertainty should be indicative of the prediction
possibly being far from the ground truth.

The results of this evaluation are displayed in Table~\ref{tab:pearson}.
Our model is leading the evaluation for the CALFIN and Baumhoer datasets,
reaching respectable Pearson coefficients of 0.4811 and 0.6031, respectively.
On the TUD dataset, HED-UNet and the classic Deep Snake
outperform our model on the uncertainty benchmark,
achieving Pearson coefficients of 0.5518 and 0.4959.
Still, our model scores decently with a Pearson coefficient of 0.4414.

These observations support our hypothesis that the contour representation is better 
suited for uncertainty quantification.

\begin{table}
  \centering
  \caption{
      Results of the Loss Functions Ablation Study.
      Deviations calculated using the Polis Metric.
  }\label{tab:losses}
  \vspace{-12pt}
  \include{ablations/losses}
\end{table}

\subsection{Ablation Studies}
In order to better understand how the design decisions help
our proposed model to improve upon the existing contour-based methods,
we conduct a number of ablation studies to quantify the value of
the network's components.

\subsubsection{Loss Function}
The rationale for implementing SoftDTW loss for our model was the assumption
that the model cannot always correctly guess the placement of the vertices
along the ground truth contour.
Indeed, we observe better performance when using a time-warping loss,
as can be seen in Table~\ref{tab:losses}.
Surprisingly, the difference between DTW and its smooth SoftDTW variant
is rather small,
which suggests that the theoretical advantage of SoftDTW's smoothness
does not matter much in practice for this application.

\subsubsection{Number of Vertices}\label{sec:vertices}
\begin{table}
  \centering
  \caption{
      Results of the Study on the Number of Vertices.
      Deviations calculated using the Polis Metric.
  }\label{tab:vertices}
  \vspace{-12pt}
  \include{ablations/vertices}
\end{table}
    When choosing the number of vertices to represent the contours,
    a balance needs to be taken between too few vertices and
    too many vertices.
    Too low a number of vertices will not allow the model to sufficiently approximate the true
    contour,
    while too many vertices should lead to overfitting and issues in communicating
    information between vertices far apart in the sequence.
    In order to experimentally find a good setting for the number
    of vertices,
    we train COBRA configurations with different numbers of vertices.
    For computational efficiency,
    we always set the number of vertices as a power of two,
    choosing 16, 32, 64, 128 and 256 as possible vertex counts.
    The results of these experiments are displayed in Table~\ref{tab:vertices}.
    
    For all three datasets, we observe that with increasing vertex count,
    performance decreases towards both ends of the tested range,
    which suggests that there is indeed a sweet spot around the middle of the
    evaluated range.
    For the CALFIN dataset, there appears to be an optimal performance
    plateau from 32 to 128 vertices
    while for the TUD dataset, 64 vertices is optimal.
    Interestingly the Baumhoer dataset seems to require a higher number of vertices for
    the best performance, reaching the best performance at 128 vertices.
    We attribute this to the aforementioned higher complexity of calving fronts
    in Antarctica. 
    In practice, we recommend to choose the number of vertices accordingly to the complexity of the calving fronts of the region of interests. In general, setting it to 64 offers overall good performance across all study areas concerned in this study, the selection of which could be quite representative for large scale applications.

\subsubsection{Number of Iterations}
\begin{table}
  \centering
  \caption{
      Results of the Study on the Number of Iterations.
      Deviations calculated using the Polis Metric.
  }\label{tab:iterations}
  \vspace{-12pt}
  \include{ablations/iterations}
\end{table}

A fundamental hyperparameter of our network
is the number of iterations of the snake head.
When given too few iterations,
the model will likely not have enough
capacity to converge to the right contour.
On the other hand, given a large number of iterations,
we expect the model to overfit on the training set
and generalize worse to unseen scenes.
In order to find evidence for these hypotheses,
we conduct a study on the number of iterations
where we retrain COBRA models
with iteration numbers from two to seven.
The results in Table~\ref{tab:iterations} suggest
that there is indeed a sweet spot at four iterations.
Starting from two iterations, performance improves
considerably on all evaluation datasets up until four iterations.
After that, increasing the number of iterations
decreases the performance again.
Therefore, we set the number of iterations for our model
to four.

\subsubsection{Coordinate Features}
Including the vertex coordinates as additional features
allows the Snake Head to take the distance and relative position
of the vertices into account, but could also introduce a source of overfitting.
In the ablation study (Table~\ref{tab:ablation})
we observe that these coordinate features improve performance slightly
on the CALFIN test set and drastically improve performance
on the TUD dataset, where the average deviation is more than halved.
For the Baumhoer dataset, performance degrades slightly
when including coordinate features.
This suggests that the coordinate features
help the model to learn implicit shape priors
for Greenlandic glacier calving fronts,
which are not helpful when transferring
the model to the Antarctic calving fronts in the Baumhoer dataset.

\begin{table}
  \centering
  \caption{
      Results of the Binary Ablation Study.
      Deviations calculated using the Polis Metric.
  }\label{tab:ablation}
  \vspace{-12pt}
  \include{ablations/binary}
\end{table}

\subsubsection{Gradient Stopping}
Originally, the idea of stopping the gradients from flowing through the vertex coordinates
between iterations was introduced to improve convergence of the model.
However, the ``No Gradient Stopping''
ablation in Table~\ref{tab:ablation} shows that this choice is essential
for the performance of the model.
Without gradient stopping, the model predictions deteriorate to
a degree where they are worse than the predictions of the baseline UNet model.
We attribute this to numerical instabilities in the
texture sampling procedure that is used to translate between the feature maps
and vertex features,
which can arise from letting gradients flow through the vertex positions.

\subsubsection{Deep Supervision}
During training, we calculate a loss term after each
snake iteration and sum up these individual loss terms
for the final loss.
To quantify the contribution of this deep supervision,
we also evaluate a model trained without intermediate loss terms,
displayed as ``No Deep Supervision'' in Table~\ref{tab:ablation}.
While the in-distribution samples from the CALFIN test set
do not improve much with deep supervision,
generalization on TUD and Baumhoer is improved
by this change.
\subsubsection{Weight Sharing}
The underlying hypothesis for shared weights in the Snake Head iterations
was the assumption that a single set of weights would lead to better
generalization results than applying a series of multiple distinct Snake Heads.
The ablation results for ``No Shared Weights'' in Table~\ref{tab:ablation}
support this hypothesis.
On the CALFIN test set,
the prediction accuracy is nearly constant between the model with
shared weights and the one with distinct weights.
However, on the other test sets,
the performance improves considerably when sharing the weights between the
Snake Head iterations.

\section{Conclusion}
We proposed an approach to detecting calving fronts
that directly predicts the desired contours
instead of predicting dense masks as an intermediate output.
By training our method and existing methods on the CALFIN dataset,
we showed that this new approach outperforms previous methods
both on the CALFIN and TUD test sets,
and exhibits competitive performance on the Baumhoer test set.
In our ablation study,
we showed the importance of network elements like
the loss function, stopping gradient flow in the Snake Head,
and sharing the weights between iterations.

Further, we showed that deep active contour models
not only provide accurate delineations of calving fronts,
but also are naturally suited for the quantification
of the prediction uncertainties.

We hope that this study can inspire new ways
of approaching similar tasks in remote sensing
where boundaries are studied,
like grounding line detection or firn line detection.

Finally, we believe that the shift in representation from
pixe-lwise masks to GIS-native data structures like polylines 
will not only reduce the computational burden,
but also allow for exciting new approaches
like enforcing of physical constraints and temporal consistency or
analysis across different coordinate reference systems.


\bibliographystyle{IEEEtran}
\bibliography{IEEEfull,refs}

\end{document}

%% file: results.tex
  \begin{tabular}{lrrrr}
    \toprule
    & CALFIN & TUD & Baumhoer \\
    \midrule
    UNet\cite{ronneberger2015_unet}               & 224 m & 288 m & 122 m \\
    DeepUNet\cite{loebel_extracting}              & 138 m & 241 m & 92 m \\
    HED-UNet\cite{heidler2022_hedunet}            & 130 m & 231 m & 122 m \\
    CALFIN\cite{cheng2021_calving}           & 138 m & 159 m & 99 m \\
    Deep Active Contours\cite{rupprecht2016_deep} & 515 m & 699 m & 652 m \\
    Deep Snake RN50\cite{peng2020_deep}           & 289 m & 467 m & 247 m \\
    Deep Snake Xception\cite{peng2020_deep}       & 123 m & 316 m & 130 m \\
    DANCE RN50\cite{liu2021_dance}                & 118 m & 290 m & 131 m \\
    DANCE Xception\cite{liu2021_dance}            & 103 m & 272 m & 102 m \\
    COBRA (mean of 3)                        &  99 m & 144 m &  99 m \\
    \midrule
    COBRA (standard deviation) & $\pm$10 m &  $\pm$ 21m& $\pm$12m\\
    \bottomrule
  \end{tabular}

%% file: pearson.tex
\begin{tabular}{lrrr}
\toprule
Dataset & CALFIN & TUD & Baumhoer \\
\midrule
UNet\cite{ronneberger2015_unet}                & 0.3603 & 0.3864 & 0.1873 \\
DeepUNet\cite{loebel_extracting}               & 0.2762 & 0.3726 & 0.4191 \\
HED-UNet\cite{heidler2022_hedunet}             & 0.3391 & \textbf{0.5518} & 0.3989 \\
CALFIN\cite{cheng2021_calving}                    & 0.1790 & 0.3449 & 0.2337 \\
Deep Active Contours\cite{rupprecht2016_deep}  & 0.0842 & 0.2190 & 0.1759 \\
Deep Snake Xception\cite{peng2020_deep}        & 0.2009 & 0.4959 & 0.4508 \\
DANCE Xception\cite{liu2021_dance}             & 0.2972 & 0.3346 & 0.3589 \\
COBRA                                          & \textbf{0.4811} & 0.4414 & \textbf{0.6031} \\
\bottomrule
\end{tabular}

%% file: ablations/losses.tex
\begin{tabular}{lrrrr}
\toprule
Loss Function & CALFIN & TUD & Baumhoer \\
\midrule
$L_2$         & 114 m & 312 m & 119 m \\
$L_1$         & 102 m & 296 m & 111 m \\
DTW           & 90 m & 236 m & 91 m \\
SoftDTW       & 99 m & 144 m & 99 m \\
\bottomrule
\end{tabular}

%% file: ablations/vertices.tex
\begin{tabular}{lrrrrrr}
\toprule
Vertices & 16  &    32 &    64 &  128 &  256 \\
\midrule
CALFIN   & 120 m &  98 m &  99 m &  98 m & 116 m \\
TUD      & 252 m & 209 m & 144 m & 214 m & 261 m \\
Baumhoer & 128 m & 102 m &  99 m &  85 m & 128 m \\
\bottomrule
\end{tabular}

%% file: ablations/iterations.tex
\begin{tabular}{lrrrrrr}
\toprule
Iterations & 2 & 3 & 4 & 5 & 6 & 7 \\
\midrule
CALFIN & 193 m & 147 m & 99 m & 157 m & 152 m & 157 m \\
TUD & 318 m & 262 m & 144 m & 271 m & 267 m & 278 m \\
Baumhoer & 148 m & 119 m & 99 m & 110 m & 111 m & 106 m \\
\bottomrule
\end{tabular}

%% file: ablations/binary.tex
\begin{tabular}{lrrrr}
\toprule
Ablation & CALFIN & TUD & Baumhoer \\
\midrule
Full Model & 99 m & 144 m & 99 m \\
\midrule
No Coordinate Features & 95 m & 301 m & 100 m \\
\midrule
No Gradient Stopping & 397 m & 285 m & 531 m \\
\midrule
No Deep Supervision & 102 m & 207 m & 118 m \\
\midrule
No Shared Weights & 88 m & 294 m & 141 m \\
\bottomrule
\end{tabular}